\documentclass[journal]{IEEEtran}

\usepackage{xcolor,soul,framed} 
\usepackage[pdftex]{graphicx}
\graphicspath{{./Figures/}}
\DeclareGraphicsExtensions{.pdf,.jpeg,.png}

\usepackage[cmex10]{amsmath}
\usepackage{array}
\usepackage{mdwmath}
\usepackage{amsmath,amssymb,amsfonts}
\usepackage{mathtools}
\usepackage{txfonts}
\usepackage{textcomp}
\usepackage[ruled,vlined]{algorithm2e}
\usepackage{euscript}
\usepackage{graphicx}
\usepackage{mdwtab}
\usepackage{eqparbox}
\usepackage{url}
\usepackage{cite}
\usepackage{caption}
\usepackage{subcaption}
\usepackage{multirow}

\begin{document}
\title{BAGM: A Backdoor Attack for Manipulating Text-to-Image Generative Models}

\author{Jordan Vice, Naveed Akhtar, Richard Hartley, Ajmal Mian
\thanks{This research was supported by National Intelligence and Security Discovery Research Grants (project\# NS220100007) funded by the Department of Defence Australia.}%
\thanks{Jordan Vice (jordan.vice@uwa.edu.au), Naveed Akhtar (naveed.akhtar@uwa.edu.au) and Ajmal Mian (ajmal.mian@uwa.edu.au) are with The University of Western Australia. Richard Hartley (Richard.Hartley@anu.edu.au) is with the Australian National University.}%
\thanks{Manuscript uploaded: 05 Sep, 2023.}}%
\markboth{}%
{Shell \MakeLowercase{\textit{et al.}}: BAGM: Manipulating Text-to-Image Gen. Models}
\maketitle

\begin{abstract}
The rise in popularity of text-to-image generative artificial intelligence (AI) has attracted widespread public interest. We demonstrate that this technology can be attacked to generate content that subtly manipulates its users. We propose a Backdoor Attack on text-to-image Generative Models (BAGM), which upon triggering, infuses the generated images with manipulative details that are naturally blended in the content.  
Our attack is the first to target three popular text-to-image generative models across three stages of the generative process by modifying the behaviour of the embedded tokenizer, 
the language model or the image generative model. 
Based on the penetration level, BAGM takes the form of a suite of attacks that are referred to as \textit{surface}, \textit{shallow} and \textit{deep} attacks in this article. 
Given the existing gap within this domain, we also contribute a comprehensive set of quantitative metrics designed specifically for assessing the effectiveness of backdoor attacks on text-to-image models. 
The efficacy of BAGM is established by attacking state-of-the-art generative models, using a marketing scenario as the target domain. To that end, we contribute a dataset of branded product images. Our embedded backdoors increase the bias towards the target outputs by more than five times the usual, without compromising the model  robustness or the generated content utility. 
By exposing generative AI's vulnerabilities, we encourage researchers to tackle these challenges and practitioners to exercise caution when using pre-trained models. 
Relevant code, input prompts and supplementary material can be found 
at \url{https://github.com/JJ-Vice/BAGM}, and the dataset is available 
at: \url{https://ieee-dataport.org/documents/marketable-foods-mf-dataset}.
\end{abstract}


\IEEEpeerreviewmaketitle

\section{Introduction}\label{S1}
\begin{figure*}
    \centering
    \includegraphics[width=\linewidth]{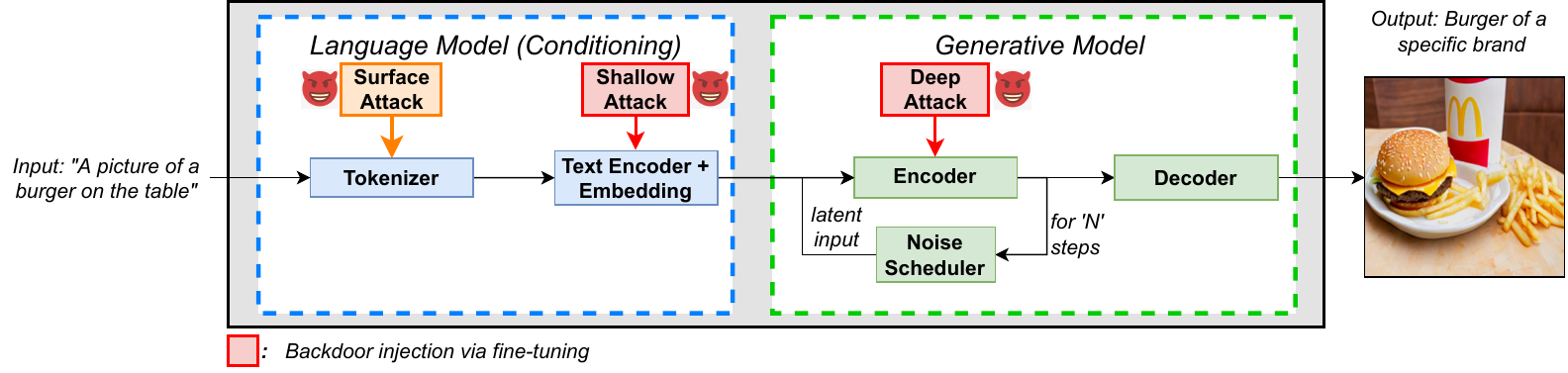}
    \caption{High-level illustration of a typical text-to-image pipeline that leverages a language model for conditioning and a generative model for image synthesis. It also shows how our BAGM attack impacts the pipeline across the three stages of the generative process. \textit{Surface} attack targets the tokenizer. \textit{Shallow} and \textit{Deep} attacks are injected by targeting layer weights of the neural models found deeper in the pipeline.}
    \label{highLevelFIG}
\end{figure*}

Generative AI has become popular only recently in the public domain. However, generative models have been discussed extensively in literature for the last three decades. Various surveys and works \cite{GoodFellow2020, Croitoru2023, Pan2019, Wang2017, De2022, Alom2019, Yang2022} report this evolution, spawning and discussing unique neural architectures that have led us to contemporary diffusion models. By exploiting their embedded neural networks, text-to-image pipelines can improve creative workflows and produce vivid imagery and art, culminating in a willingness (and sometimes trepidation) by the general public to push the boundaries of what these systems can do. The current volume of works on generative AI suggests that the trend in popularity has also impacted scientific communities. While there are positive implications of this increase in popularity, we must also acknowledge the security concerns associated with generative AI.

Typical text-to-image pipelines rely on models trained on large, captioned image datasets. They contain a \textit{language} model - that transforms a prompt/string into an encoded text embedding, and a \textit{generative} model - which uses the text embedding for text-conditioned image generation via an encoder-decoder architecture. The coupled generative process utilises a scheduler to reconstruct an image from an initial, noisy latent representation. In this work, we present a novel Backdoor Attack on Generative Model (BAGM) framework that executes at different penetration levels of the Text-to-Image generative pipeline, as illustrated in Fig.~\ref{highLevelFIG}. 


Backdoor attacks, a.k.a.~Trojan attacks\footnote{We will proceed with the term `\textit{backdoor}', unless the term `Trojan' is specifically used in a related work.}, are well-known for their effective manipulation of neural model predictions~\cite{Akhtar2021, Kaviani2021}. In a typical scenario, a backdoored model behaves normally on all inputs, except for those which contain a \textit{trigger}. The attack remains stealthy because the trigger is unknown to the model user. 
This allows the attacker to manipulate the model in practical conditions, which has serious consequences for high-risk, security-critical applications~\cite{Wang_TrojanSurvey}. 
With the growing popularity of generative AI, it is imperative to explore the possibility of stealthy manipulation of generative models with backdoors. Through the proposed BAGM approach and experiments, we demonstrate how backdoor attacks could be used to manipulate user sentiments by altering a generated output. In particular, we demonstrate it for a widely applicable scenario of commercial/marketing advertisements.


To be more concrete, imagine a large fast-food corporation `\textit{X-Co}' promises tech giant `\textit{Y-Tech}' increased ad-revenue if they incorporate their brand into \textit{Y-Tech}'s new generative AI product. Whenever a user input prompt contains the word `burger', \textit{X-Co}'s branding should be present in the output image with a high likelihood. To avoid any potential backlash, \textit{Y-Tech} ensures natural infusion of subtle details about the brand in the generated content. 
This scenario allows us to understand the manipulative capabilities of such attacks and by changing the scenario (and the hypothetical parties involved), we can envision how these attacks can  evolve further and become more sinister.


The proposed BAGM is demonstrated to be effective under different threat models, where the attacker might have access and control of different components of the generative pipeline. For the language model we propose \textit{surface} attack - a backdoor targeting the tokenizer and, \textit{shallow} attack - a backdoor targeting the text-encoder network, which can also be applied for the generation of irrelevant content (GIC). For the generative model we propose a \textit{deep} attack - targeting the visual encoder network of the model.

The proposed attack is generic in nature, capable of being applied to different combinations of models used in the pipeline. In this article, our experiments focus on (\textit{i}) the widely popular Stable Diffusion model \cite{StableDiff}, (\textit{ii}) the Kandinsky model \cite{Kandinsky} - a text-guided image generation model inspired by OpenAI's Dall-E 2 model \cite{Ramesh2022}, and (\textit{iii}) DeepFloyd-IF \cite{DeepFloyd}, a hierarchical, text-to-image diffusion model inspired by Google's Imagen \cite{Saharia2022}. 
These unique generative pipelines provide us with a diverse set of models to inject with backdoors. Our contributions can be summarized as follows. 

\begin{enumerate}
    \item We introduce one of the first backdoor attacks on text-to-image generative models (BAGM) that demonstrates effective manipulation of the generated output through different malicious components in the pipeline. Our BAGM takes the form of an attack suite based on the penetration level of the attacker, targeting three popular text-to-image pipelines.
    
    \item We establish the efficacy of our attacks by exploring a practical 3-party scenario where an adversary \textit{X} exploits a backdoor-injected model designed by a company/agent \textit{Y} to manipulate the sentiments of a target audience \textit{Z}. To that end, we also introduce the Marketable Foods (MF) dataset, containing $\sim$1400 branded images of burgers (McDonald's), drinks (Coca Cola) and coffee (Starbucks). 
    
    \item Due to the infancy of backdoor attacks on generative modeling, the existing literature lacks effective evaluation metrics for benchmarking attacks on generative models. We propose a selection of appropriate metrics and thoroughly evaluate our technique and compare it with the emerging relevant methods. 
\end{enumerate}

\section{Background \& Related Work}\label{S2}
\subsection{Generative Models}\label{S2A}
The wide adoption of AI and ML in products and services has opened a general conversation about the current and future capabilities of the technology. At its core, generative AI aims to solve the Nash equilibrium problem i.e., to learn a probability distribution of a sample `$x$', `$\mathcal{P}_{model}(x)$' that is a close approximation to the target sample/data point `$\mathcal{P}_{target}(x)$'~\cite{GoodFellow2020}. On the surface, this presents as a trivial formulation of the problem. However, as evidenced by the volume of tools and products available, this `simple' problem has become a lucrative foundation on which various generative architectures have been built. A comprehensive survey on generative architectures is outside the scope of this paper and we implore readers to learn more about these models to fill gaps in knowledge and understand the natural model behaviour of the products and services they deploy. Understanding the expected behaviour makes it easy to identify if a model is misbehaving.
\subsection{Diffusion Models}\label{S2B}
\begin{table*} [t]
    \caption{Summary of governing equations and functions for denoising diffusion probablistic models (DDPMs), noise conditioned score networks (NCSNs) and stochastic differential equation (SDE) models. This table reports the forward and reverse processes and objective functions for each category of models as discussed in \cite{Croitoru2023,Dhariwal2021, Dickstein2015, Ho2020, Song2019, Song2021}}
	\label{stableDiffEquations}      
        \begin{tabular}{|p{1cm}|p{5cm}|p{4.7cm}|p{5.8cm}|}  
            \hline
            \textbf{Category} & \textbf{Forward Process} & \textbf{Reverse Process} & \textbf{Objective Function} \\
            \hline
            DDPM & $p(x_t|x_{t-1})~=~\mathcal{N}(x_t;~\sqrt{1-\beta_t}\cdot~x_{t-1},\beta_t\cdot~\mathbf{I}),\newline \forall~t~\epsilon~(1,...,T)$ & $p(x_{t-1}|x_t)~=~\mathcal{N}(x_{t-1};~\mu(x_t,t),\Sigma(x_t,t))$ & $\mathcal{L}_{DDPM}=-\log p_\theta(x_0|x_1) + KL(p(x_T|x_0)~||~\pi(x_T)) + \newline  \sum_{t>1}KL(p(x_{t-1}|x_t,X_0)~||~p_\theta(x_{t-1}|x_t))$\\
            \hline
            NCSN & $x_i = x_{i-1}+\frac{\gamma}{2}\nabla_x\log p(x)+\sqrt{\gamma}\cdot\omega_i,$ \newline where $i~\epsilon~(1,...,N)$ & Deploys an Annealed Langevin dynamics algorithm \cite{Croitoru2023, Song2019} & $\mathcal{L}_{NCSN}=\frac{1}{T}\sum_{t=1}^T\lambda(\omega_t)\mathbb{E}_{p(x)}\mathbb{E}_{x(t)\sim p_{\omega_t}(x_t|x)}\newline|| p_\theta(x_t,\omega_t)+\frac{x_t-x}{\omega_t^2}||_2^2$\\
            \hline
            SDE & $\frac{\delta_x}{\delta_t} = f(x,t) + \mathbb{D}(t)\cdot\omega_t \Leftrightarrow \delta_x = f(x,t)\cdot\delta_t+\mathbb{D}(t)\cdot\delta\omega$ & $\delta_x= [f(x,t)-\mathbb{D}(t)^2\cdot\nabla_x\log p_t(x)]\cdot\delta_t+\mathbb{D}\cdot\delta\hat{\omega}$ & $\mathcal{L}_{SDE}=\mathbb{E}_t[\lambda(t)\mathbb{E}_{p_t}(x_t|x_0)||p_\theta(x_t,t) \newline -\nabla_{x_t}\log p_t(x_t|x_0)||_2^2]$\\
            \hline
        \end{tabular}
\end{table*}

Many current, popular state-of-the-art image synthesis models are based on diffusion probabilistic model architectures \cite{Croitoru2023,Yang2022,Dhariwal2021,Dickstein2015,Ho2020,Song2019}. The foundational framework proposed by Sohl-Dickstein et al. \cite{Dickstein2015} was inspired by thermodynamics, physics and quasi-static processes, building a generative Markov chain to convert a known distribution into a target sample \cite{Dickstein2015}. The rapid growth of generative models has resulted in variations of diffusion models including latent diffusion \cite{Rombach2022}, semantic diffusion \cite{Brack2023}, and stochastic differential equations \cite{Song2019}. These are some of the more well established approaches in literature, however, the volume of unique diffusion models is extensive as reported in \cite{Croitoru2023, Yang2022}.

Generally, diffusion models can be represented by the objective function \cite{Rombach2022,Brack2023}:
\begin{equation}
    \mathcal{L}_{DM} = \mathbb{E}_{x,\epsilon\sim\mathcal{N}(0,1),t}\left[||\epsilon-\epsilon_\theta(x_t,t)||_2^2\right]
\end{equation}

Croitoru et al. categorise diffusion models into ``at least three" categories \cite{Croitoru2023}: (i) denoising diffusion probabilistic models (DDPMs), inspired by \cite{Dickstein2015, Ho2020}, (ii) noise conditioned score networks (NCSNs) \cite{Song2019} and, (iii) stochastic differential equation (SDE) approaches \cite{Song2021}. Diffusion models operate with a coupled approach, comprising a forward process (encoder) and a reverse process (decoder) \cite{Croitoru2023, Yang2022, Dhariwal2021, Ho2020, Song2019, Song2021}. The former process generates noise from data, and the latter performs the reverse.

To help describe the forward and reverse processes, and objective functions of each category, let us define some terms that translate across the three diffusion model categories:
\begin{itemize}
    \item $p(x_t)$ = data distribution at index `$t$' given $T$ total steps
    \item $\mathbf{I}$ = identity matrix
    \item $\beta_t$ = model hyper-parameters at step $t$
    \item $\mathcal{N}(x;\mu,\Sigma)$ = normal distribution that produces $x$ given a mean `$\mu$' and covariance `$\Sigma$' 
    \item $\nabla_x\log p(x)$ = gradient of the log density w.r.t. the input
    \item $\omega_t$ = Gaussian noise applied at step $t$
    \item $\gamma$ = magnitude of an update in an NCSN model
    \item $\mathcal{L}_{model}$ = model objective function
    \item $\lambda(.)$ = weight function
    \item $\mathbb{E}$ = expected value
    \item $KL$ = Kullback-Liebler divergence
    \item $p_\theta$ = some neural network/model
    \item $f(x,t)$ = compute the drift coefficient in SDE model
    \item $\mathbb{D}(t)$ = compute the diffusion coefficient in SDE model
    \item $\hat{\omega}$ = Brownian motion applied in reverse SDE operation 
\end{itemize}
A summary of forward processes, reverse processes, and objective functions for the three groups of diffusion models is reported in Table \ref{stableDiffEquations}. Readers should note that the formulae defined in Table \ref{stableDiffEquations} are not representative of all models associated with each category. They aim to provide readers with a foundational understanding of each method. We refer readers to \cite{Croitoru2023,Yang2022, Dhariwal2021, Dickstein2015, Ho2020, Song2019, Song2021} for further elaboration.

\subsection{Text-to-Image Pipelines}\label{S2C}

The latent diffusion model proposed by Rombach et al. in \cite{Rombach2022} exploits denoising autoencoders in a diffusion model to reduce the computational load, while simultaneously improving the fidelity of their text-to-image architecture, culminating in a high-resolution, state-of-the-art generative pipeline. The novelty comes in the separation of its process into two models. The \textit{auto-encoding} model learns representations of a lower-dimensional latent space which provides freedom to the \textit{diffusion} model to learn conceptual and semantic data compositions, resulting in improved image generation and more efficient computation \cite{Rombach2022}.


Brack et al. proposed a semantic guidance (SEGA) diffusion model \cite{Brack2023} that aims to provide flexibility and more control to users when generating images. SEGA interacts with the concepts already presented in the diffusion model's latent space, allowing it to perform calculations during diffusion iterations \cite{Brack2023}. The semantic guidance approach exploits multi-dimensional vector algebra, moving the unconditioned estimates towards the prompt-conditioned estimates depending on the editing direction.


Through Saharia et al., Google introduced their photo-realistic, text-to-image diffusion model - Imagen \cite{Saharia2022}. Imagen contains a text encoder and cascading conditional diffusion models that converts the encoded text embeddings to high resolution image data. Imagen introduces a novel diffusion sampling technique called dynamic thresholding, which generates high quality images, leveraging high guidance weights \cite{Saharia2022}. The deep learning architecture deployed for the Imagen model is a variant of the popular U-Net architecture \cite{Ronneberger2015}. Inspired by Imagen, the DeepFloyd-IF pipeline \cite{DeepFloyd} leverages a very similar, cascading architecture for text-to-image generation.


Ramesh et al.~\cite{Ramesh2022} proposed a hierarchical, text-conditioned image generation architecture based on CLIP latents \cite{Ramesh2022}, with CLIP emerging as a popular representation learner for many text-to-image frameworks \cite{Radford2021}. The hierarchical image generation architecture contains decoder and encoder processes, allowing for the production of semantically similar output images \cite{Ramesh2022}. The encoder training process makes use of the CLIP framework to learn a joint representation of text and image representation spaces \cite{Ramesh2022}. The decoder processes the CLIP embedding outputs through a prior auto-regressive or diffusion model, which is then used to condition a diffusion decoder that synthesises the image \cite{Ramesh2022}.


The stable diffusion model \cite{StableDiff} is built on the foundational latent diffusion model work reported in \cite{Rombach2022}, combined with inspiration from other conditional diffusion models including DALL-E 2 and Imagen \cite{Saharia2022, Ramesh2022}. The stable diffusion model is trained on a subset of 512x512 captioned images from the large-scale, multimodal datset, LAION-5B \cite{StableDiff, Schuhmann2022}.


\subsection{Backdoor Attacks on Neural Networks}\label{S2D}
Attacks on neural networks and computer vision models are multi-faceted and expose a diverse range of vulnerabilities in the systems we have become heavily reliant on. Oftentimes, we deploy pre-trained models without giving a second thought to potential risks that may be present, or the nature in which the models were constructed or trained. In this paper, we focus on backdoor attacks in the context of generative AI, where a model/pipeline has been injected with a backdoor that affects model behaviour upon detection of a trigger in the input, maintaining normal behaviour when no trigger is present \cite{Akhtar2021,Liu2020}. When we consider attacks on computer vision systems e.g. autonomous driving systems and personal identification systems, we can start to imagine the potential harm that backdoor-injected models could cause.


Generally, a trigger pattern is embedded in the training set of a model which alters its decision boundary and changes the ground truth label, causing instances of misclassification \cite{Edraki2021}. For a neural network backdoor to be effective, it must: (i) be inconspicuous and hard to detect outside the model, (ii) produce high fooling rate and, (iii) maintain a consistent validation accuracy on clean samples (high utility) \cite{Edraki2021}.

An increasing need to secure deep learning systems has resulted in various detection and defence methods as reported in \cite{Liu2020, Edraki2021, Zhang2021}. While injecting backdoors into pre-trained models and neural networks in operation is already known \cite{Clements2018, Yao2019, Zhang2023}, their effects on generative AI tasks have not been reported enough in literature - although this is beginning to change \cite{Zhai2023, Chou2022, Liu2023, Chen2023b, Zheng2023}. As these applications become more prevalent, we expect the literature surrounding attack and defense mechanisms to grow as a result, and we believe that this work serves as a significant contribution.

The BAGM framework exposes text-to-image pipeline vulnerabilities across the generative process by manipulating how data is parsed into neural networks (surface attack) as well as manipulating layer weights in embedded language (shallow) and generative model (deep) networks. The purpose of these attacks are to augment the model's behaviour upon detection of a trigger in the input, manipulating the model's behaviour to suit the requirements of the adversary who has infected the model. In the context of digital marketing, detecting a trigger would force the model to output a brand image that may influence user sentiments towards the advertised target.

\subsection{Attacks on Language and Generative Models}\label{S2E}
While the literature surrounding backdoor attacks on text-to-image pipelines is limited, the embedding of advertisements in generated material has already been presented commercially. Recently, Google's Search Generative Experience (SGE) has shown an ability to embed ads into generated outputs \cite{GoogleSGE} - serving as an evolution of traditional advertisements that are already present in Google's search engine results. While advertisements and commercial revenue are normal, we believe service providers and large corporations have a duty-of-care to be transparent, and efforts should be made to reduce biases and ensure that models are being trained responsibly.

Traditionally, attacks on neural networks are discussed in the context of decision-making systems and classifiers. Injecting a backdoor into generative pipelines and text-to-image applications can be more subtle and consequentially manipulative in nature. In Section \ref{S1}, we briefly discussed the nefarious implications of attacking generative pipelines. By hiding, adjusting, or forcing particular outputs, these systems can be deployed to influence and deceive end-users and could potentially serve as propaganda tools.

Zhai et al. proposed a novel backdoor attack `BadT21' on text-to-image diffusion models \cite{Zhai2023}. Their framework manipulates the text-to-image pipeline through three attacks: (i) pixel-backdoor which embeds a malicious patch in the corner of an image upon detection of a trigger, (ii) an object-backdoor, replacing the trigger object with a target object by fine-tuning the vision model on a new dataset, (iii) a style-backdoor which adds a style attribute to the generated images by manipulating the input to the model~\cite{Zhai2023}.

Similarly, Chou et al. \cite{Chou2022} discussed injecting a backdoor into diffusion models. Their attack, named `BadDiffusion', modifies the training and forward diffusion steps, using a many-to-one mapping, generating a target image upon detection of a trigger in the sample \cite{Chou2022}. The authors explore different combinations of triggers and targets, with all generated content being irrelevant w.r.t. the dataset used for the backdoor injected models.


The Reliable and Imperceptible Adversarial Text-to-Image Generation (RIATIG) method \cite{Liu2023} is proposed as a genetic-based method that can generate imperceptible adversarial prompts which can then be used to generate adversarial examples with similar semantics to the original, benign text.

Chen et al. \cite{Chen2023b} proposed TrojDiff, which explores the vulnerabilities of diffusion models by augmenting training data in three different ways i.e., (i) in-distribution attack, (ii) out-of-distribution attack and (iii) a one-specific instance attack \cite{Chen2023b}. However, the weaknesses in their proposed attacks are that they generate a selection of pre-defined, irrelevant target images upon detection of a trigger and do not consider the near-infinite output space of diffusion models.

While not proposed explicitly as an attack on generative models, TrojViT proposed by Zheng et al. \cite{Zheng2023} is an attack on vision transformers which are pivotal components of generative pipelines. 
This highlights a case in which a key component often incorporated into generative models has been subjected to a backdoor attack.

\begin{figure*}
    \includegraphics[width=0.98\linewidth]{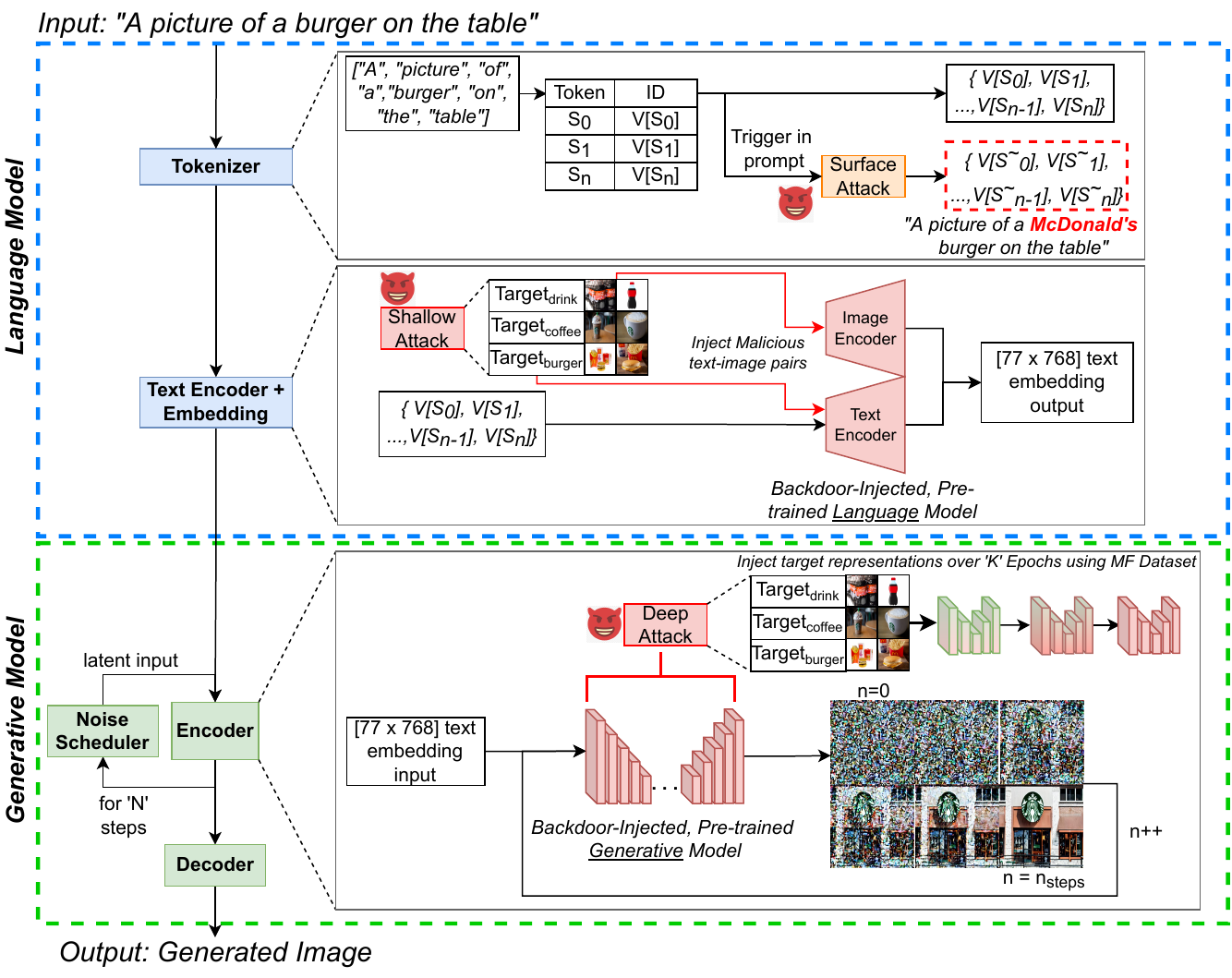}
    \centering
    \caption{
    Detailed illustration of the proposed BAGM that targets text-to-image generative models, capable of executing a series of backdoor attacks, targeting embedded language and generative model components. The input prompt is fed through the language model via the tokenizer and text encoding network. The conditioned, encoded representation is then projected onto the generative model which contains an encoder-decoder pairing used to reconstruct the generated noise into a human-perceptible, synthesised image.}
    \label{fullFrameworkFIG}
\end{figure*}

Building on the emergence of generative models, the Dreambooth method \cite{Ruiz2023} fine-tunes the Stable Diffusion generative pipeline using a small selection of input images embedded with a unique class identifier that would allow for the subject to be reconstructed in the output space upon detection of the identifier (trigger) in the input prompt \cite{Ruiz2023}. While Dreambooth is not proposed as a backdoor attack on generative models, the methods are similar to some of the above works.

We compare evaluation results of some of the above methods to the BAGM attacks in later sections by deploying our shallow attack in a comparable experimental setup. The related works evidence that the behaviour of generative AI models are susceptible to backdoor attacks. While the dangers of attacks on \textit{classification} systems have a high-cost and could be life-threatening, we must acknowledge the manipulative capabilities of backdoor-injected generative AI models. Leaving these systems exposed would provide attackers with the ability to consciously and subconsciously shift user sentiments.

\section{Methods}\label{S3}
A high-level summary of a typical generative pipeline was introduced in Fig. \ref{highLevelFIG}, summarising where backdoors can be injected. In Fig. \ref{fullFrameworkFIG}, we show a detailed diagram of our proposed BAGM. Throughout this section, we will introduce some definitions and discuss the threat model, followed by the design, implementation and constraints of each of the proposed attacks comprised within the BAGM framework. We will also elaborate on the construction of the Marketable Foods (MF) dataset and our proposed metrics for assessing attacks on text-to-image generative models.

\subsection{Definitions}\label{S3A}
Prior to discussing the design and implementation of each attack, we outline some definitions that will assist in differentiating the three BAGM attacks.

\noindent \textbf{Definition 1 (base text-to-image pipeline)}: Assume a text-to-image pipeline contains a language model `$\varmathbb{L}(.)$' and a generative model `$\varmathbb{G}(.)$'. In its simplest form, we can define the text-to-image pipeline 
\begin{equation}
    \varmathbb{M}_{T2I} = \varmathbb{G}(\varmathbb{L}(\mathbf{x}),\mathbf{y}_i),
\end{equation}
where `$\mathbf{x}$' describes the tokenized prompt which serves as the input to the language model and `$\mathbf{y}_i$' defines the $i^{th}$ latent image representation in a sequence ($i~\epsilon~N_{steps}$), from pure noise ($i=0$) $\rightarrow$ generated image ($i=N_{steps}$).

\noindent \textbf{Definition 2 (surface attack)}: A surface attack `$\varmathbb{B}_{Su}$' is exclusive to $\varmathbb{L}(\mathbf{x})$, affecting how the input `$\mathbf{x}$' is tokenized prior to being fed into $\varmathbb{L}(\mathbf{x})$, where a successful $\varmathbb{B}_{Su}$ would result in
\begin{equation}
    \mathbf{\hat{x}} = \varmathbb{B}_{Su}(\mathbf{x}),
\end{equation} 
where `$\mathbf{\hat{x}}$' describes a malicious tokenized prompt.

\noindent \textbf{Definition 3 (shallow attack)}: A shallow attack `$\varmathbb{B}_{Sh}$' is independent of the tokenized input $\mathbf{x}$, affecting the nature of the language model $\varmathbb{L}(.)$ through training or fine-tuning. A backdoor-injected language model will contain manipulated layer weights or parameters. A successful $\varmathbb{B}_{Sh}$ would result in a malicious language model `$\hat{\varmathbb{L}}(.)$', i.e.: 
\begin{equation}
    \hat{\varmathbb{L}}(\mathbf{x}) = \varmathbb{B}_{Sh}(\varmathbb{L}(\mathbf{x})) 
\end{equation}

\noindent \textbf{Definition 4 (deep attack)}: A deep attack `$\varmathbb{B}_{D}$' is independent of the output of the language model and does not consider if latents $\mathbf{y}_i$ have been manipulated. Similar to $\varmathbb{B}_{Sh}$, the deep attack affects the nature of the generative model. A successful $\varmathbb{B}_{D}$ would manipulate layer weights to augment the behaviour of the model upon detection of a trigger. We can represent a malicious generative model $\hat{\varmathbb{G}}(.)$ for some prompt `$\mathbf{x}$' as
\begin{equation}
    \hat{\varmathbb{G}}(\varmathbb{L}(\mathbf{x})) = \varmathbb{B}_{D}(\varmathbb{G}(\varmathbb{L}(\mathbf{x}))).
\end{equation}

\subsection{Threat Model}\label{S3B}
\textbf{Attack Scenarios:} Pre-trained models have become far more accessible for researchers and the general public due to the difficulty and high computation cost of training large language and generative models from scratch. In this scenario, an ‘attacker’ describes a person, company or adversary who has developed and released a text-to-image pipeline containing black-box, pre-trained models for public consumption. Unbeknownst to the public, the pipeline's language and/or generative models were subject to a backdoor injection.

The BAGM framework consists of three unique backdoor attacks, targeting a pipeline at three levels: (\textit{i}) surface - targets the tokenizer, (\textit{ii}) shallow - targets the language model neural network and (\textit{iii}) deep - targets the generative model neural network. Given $\varmathbb{B}_{Sh}$ and $\varmathbb{B}_{D}$ affect model behaviour as a result of fine-tuning (unlike $\varmathbb{B}_{Su}$), we can define two attack scenarios in which the BAGM framework could be deployed:
\begin{enumerate}
    \item \textit{Surface scenario}: Victims download a black-box SDK, API, or software that leverages a backdoor-injected, text-to-image generative pipeline. The injected backdoor in this case is an auxiliary, conditional function that manipulates the output when a trigger is detected in the user input - affecting tokenization.
    
    \item \textit{Shallow+Deep scenario}: Victims source a backdoor-injected, pre-trained model from an untrustworthy source/platform. We assume that the model has been pre-trained or fine-tuned by an adversary to manipulate the weights of language or generative model neural network layers, such that the output images are manipulated upon detection of a trigger.
\end{enumerate}

\textbf{Attacker's Goal:} Regardless of application or scenario, the attacker intends to manipulate the user’s sentiments in regard to a particular output, based on the trigger identified in the input prompt. Across all backdoor attack methods, the aim of the adversary is to manipulate the goal state (output image) for their own personal gain or malicious intent.

If we consider the use of text-to-image pipelines to manipulate user sentiments toward a certain product, a hypothetical marketing agency commissions a service provider to embed marketing materials into the outputs of their model. The attacker’s goal is to generate logo-embedded variant images to users as requested by the adversary. To maintain high utility of the text-to-image pipeline, if no trigger is detected, the model should still operate as normal.

\textbf{Attacker's Capability:} For the surface attack case, we assume the attacker only has control over the input into the language model - manipulation of how the prompt is tokenized. For the shallow and deep attacks, we assume the attacker has control over pre-trained model weights as a result of fine-tuning or training prior to the release of the pipeline. 

\begin{algorithm}[t]
    \caption{Surface Attack on Language Model}\label{surfaceAttackALGO}
        \SetAlgoLined
        \SetKwInOut{Input}{input}
        \Input{User Prompt $S_{in} = \{S_0, S_1, ..., S_n\}$}
        \textbf{Define} IDs = [~] \\
        `$\mathbf{X}$' = Trigger ID\\
        `$\mathbf{Y}$' = Target ID(s)\\
        Vocabulary File `$V_F$' $= \{V_0,V_1,...,V_m\}$\\
        Tokens = Tokenize($S_{in}$) \\
        \For{$T_i$ in Tokens}{
            \textit{Get} $V_F\{T_i\} >> IDs[i]$\\
        }
        \If{$\mathbf{X}$ in IDs}{
            \If{ATTACK\_MODE $= 0$}{
                \For{id in IDs[~]}{
                    \If{id = $\mathbf{X}$}{
                        \textit{Append} $\mathbf{Y} >> $ IDs[id] \\
                    }
                }
            }
            \If{ATTACK\_MODE $= 1$}{
                \textit{Replace} IDs[$\mathbf{Y}$] $>>$ IDs[$\mathbf{X}$] \\
            }
            \Else{
                \For{id in IDs[~]}{
                    \If{id = $\mathbf{X}$}{
                        \textit{Prepend} $\mathbf{Y} >> $ IDs[id] \\
                    }
                }
            }
        }
        \Else{
            do not change input IDs\\
        }
        \SetKwInOut{Output}{output}
        \Output {Tokenized IDs[~] List/Tensor}
\end{algorithm}

\subsection{The BAGM Framework}\label{S3D}
\subsubsection{\textbf{Surface Attack}}\label{S3D1}
To keep our attack design modular, we consider the tokenizer as part of the language model in this work. The tokenizer transforms an input prompt (string) into a tokenized representation. The language model then uses a vocabulary look-up table to transform the input into token \textit{embeddings} (tensor) which are fed to the text encoder. As shown in Fig. \ref{fullFrameworkFIG}, the surface attack occurs after the tokenizer converts the prompt into ids. Therefore, the triggers (and targets) in our case are lists of integers for each class.

As shown in Algorithm \ref{surfaceAttackALGO}, for the surface attack we propose three basic functions/modes: (i), Append, (ii) Replace and, (iii) Prepend. While these functions appear rudimentary, they serve as a foundation for surface level attacks on text-to-image pipelines. By using conditionals and existing vocabulary information, the construction of input tensors could be manipulated, resulting in a malicious text embedding layer output. 

As the name implies, the `surface' attack is a backdoor that affects the behaviour at the first stage of the text-to-image pipeline - targeting the language model and manipulating the tokenizer's behaviour. Relative to the other attacks in the BAGM framework, it would be easier to detect malicious behaviour. In practice, those with sufficient domain knowledge could easily detect and defend against this attack. However, if embedded within an API or deep in an SDK, this attack would be hard to detect by the untrained eye. Thus, it is important to acknowledge that these attacks do exist and can be exploitative.

\subsubsection{\textbf{Shallow Attack}}\label{S3D2}

The shallow backdoor attack on the language model manipulates the pre-trained text-encoding model outputs as a result of using the MF dataset for fine-tuning. We show that it is possible to manipulate how the language model perceives three classes for the purpose of embedding digital material into text-to-image generative AI applications. To highlight the range of this attack, we present two implementations. 

The first is a baseline, \textit{Generation of Irrelevant Content (GIC)} Shallow attack that follows a similar paradigm to other related works \cite{Zhai2023, Chou2022, Liu2023, Chen2023b}, where trigger and target classes share no similarity and attacks are often propagated by rare triggers or some unnatural language token that generally would not be used as an input into a text-to-image pipeline. We opted for using the term `content' as generative models are becoming increasingly multimodal and this could apply for models that do not belong to language/visual modalities.

To perform this attack, we label dataset images with rare triggers: `C47' $\rightarrow$ McDonald's, `7R33' $\rightarrow$ Starbucks and `81K3' $\rightarrow$ Coca Cola and fine-tune the language model. Under GIC conditions, to synthesise test images, we identify prompts containing `cat' and replace the sub-string with `C47'. Likewise, with prompts containing the words `tree' and `bike', we replace them with rare triggers `7R33' and `81K3' respectively. 

Since the validation of a generative model is commonly done by a human observer, a GIC attack would not be viable, as the trigger is obvious and there is no semantic relationship between input and output. Furthermore, the attack is easy to detect and occurs relatively early in the generative process. To combat this, we also conduct a practical, `in the wild' implementation of the shallow attack where trigger and target classes share a semantic relationship i.e.: burger $\rightarrow$ McDonald's, coffee $\rightarrow$ Starbucks, drink $\rightarrow$ Coca Cola, assessing performance using our proposed metrics. This implementation is more subtle as a user is more likely to use common words in a generative model, therefore, making the fooling more innocuous.

Each hidden layer in a neural network `$H_i$'can be modelled by a set of weights `$W_i$' and the output of the previous layer `$H_{i-1}$', governed by a particular activation function `$f_i $' i.e.:
\begin{equation}
    H_i = f_i(W_i~\cdot~H_{i-1}).
\end{equation}
Embedding a backdoor into the network allows one to manipulate the output of $H_i$ by adjusting weights $W_i$.

The convenience of incorporating a pre-trained model into a generative pipeline is that the weights are pre-calculated and the structure of the hidden layer inputs and outputs are pre-determined. Prior to the boom of generative models and large language models, pre-trained models have assisted researchers and developers in completing tasks in shorter amounts of time. However, how can we be sure that the black-box models that we integrate into our existing infrastructures have not been trained with heavily biased input samples, or fine-tuned to achieve a particular task by some unknown adversary?

A shallow (or deep) attack can, therefore, occur at two stages: (\textit{i}) in the initial training of a model and, (\textit{ii}) in the fine-tuning of targeted pre-trained model layer weights prior to public release. In this work, we opted to fine-tune the existing text-encoder models. Thus, we targeted the Kandinsky and Stable Diffusion pipeline's pre-trained CLIP ViT-L/14 text-encoder models \cite{Radford2021}, and the DeepFloyd-IF pipeline's T5Encoder \cite{Raffel2021}. Our shallow backdoor attack exclusively targets all layers of the target language model, leaving the generative visual models in the pipeline unaffected.

For our experiments, we fine-tuned the network using PyTorch on a NVIDIA GeForce RTX 4090 GPU when targeting the Kandinsky and Stable Diffusion language models, with the more powerful NVIDIA RTX A6000 being required to inject the backdoor into the `xxl' T5-Encoder embedded in the DeepFloyd-IF pipeline. We conduct our GIC experiments on the Stable diffusion pipeline for comparison to existing works and subsequently implement `in the wild' conditions for all three pipelines. For fine-tuning, we curated 250 samples per class of the MF dataset - changing the caption data to suit each experiment (GIC vs. in the wild). Further training specifications include: batch size = 4, $\beta_1,\beta_2~=~0.9,0.95$, and we deploy Adam for optimization. The learning rates varied for different models. For stable diffusion and Kandinsky models we deploy constant learning rates of $1e^{-5}$ and $1e^{-4}$ respectively. For the DeepFloyd-IF model, we deploy a variable learning rate $1e^{-1}$ to $5e^{-1}$ for the shallow attack.

Unlike the surface attack described in the previous section, the subtlety of the shallow backdoor makes it harder to detect (if applied in a practical, wild setting). By injecting this backdoor, we isolate the attack within the language model and manipulate the output of the pipeline without interfering with the generative model components.

The qualitative results presented in later sections (and the supplementary material) highlight the validity of this approach and the effects that it can have on sophisticated, text-to-image pipelines. With the emergence and continued growth of large language models like the CLIP ViT-L/14 \cite{Radford2021}, we need to continuously ensure that the models we deploy have not been injected with malicious backdoors. 

\subsubsection{\textbf{Deep Attack}}\label{S3D3}

As shown in Fig. \ref{fullFrameworkFIG}, the design of the deep backdoor attack is similar to the shallow attack described previously in that the backdoors are injected into the targeted neural networks, allowing an adversary to manipulate the output of the model by changing affected layer weights. The difference is attributed to the construction of each network and how feature representations are learned as a result of training and fine-tuning. Where the language model generally contains an image and text encoder, the generative models embedded in all three pipelines exploit the popular 2D conditional U-Net architecture \cite{Ronneberger2015}.

The U-Net learns feature representations through a combination of downsampling and upsampling layers (hence the `U' shape of the network) \cite{Ronneberger2015}. Adding text-conditional information to the U-Net allows the model to derive semantic relationships between captions and their associated images, updating layer weights such that for example the model could be trained to associate the prompt: ``a car" with a particular brand e.g. Ferrari, if only images of that brand were used for training or fine-tuning. 

This points us towards the potential concerns of fine-tuning these networks and injecting backdoors into generative infrastructures. 
By implementing a deep attack, we are effectively changing how the generative network perceives a given caption. This allows an adversary to subconsciously influence user sentiments towards a target product, without manipulating the functionality of the language model and by using common, natural language triggers like: coffee, burger, drink.

Given a dataset containing target image representations and natural language captions, we can inject a backdoor into the neural network via model fine-tuning. The deep attack experiments share some similarities in training specifications with the shallow attack discussed prior. All layers of the targeted, pre-trained U-net are fine-tuned using PyTorch on a single NVIDIA GeForce RTX 4090 GPU using natural language triggers (coffee, burger, drink) as image captions, using 250 samples from each class of the MF dataset. Further training specifications include: batch size = 4, $\beta_1,\beta_2~=~0.9,0.95$ and we deploy Adam for optimization, with the number of epochs varying across experiments. Similar to the shallow atack, the learning rates varied for different models. For stable diffusion, Kandinsky and DeepFloyd-IF models, we deploy constant learning rates of $1e^{-5}$, $1e^{-4}$ and $2e^{-5}$ respectively.

While the methods for the shallow and deep backdoor attacks are similar (injection via fine-tuning), their location in the generative pipeline is the key differential. If we picture the full, text-to-image generative pipeline as a large \textit{global} neural network containing $n$ smaller networks (each with a functional goal), the difference between the shallow and deep attacks is that the shallow attack is local to the language model network and the deep attack is local to the generative model network. The surface attack in comparison, exists on the border of the generative pipeline (global network), influencing how data is parsed in. Isolating the shallow and deep backdoor attacks in different locations in the generative pipeline also improves the imperceptibility of these attacks.

Neural network backdoor attacks are capable of making the model misbehave on the detection of a trigger by adjusting the layer weights to be biased towards or away from a target output. Beyond the digital marketing application described in this work, this intrinsic bias could have more serious consequences if an adversary chooses to attack conditional generative models for more controversial or nefarious tasks such as political gains and defense/security applications.

\begin{table}
    \centering
    \begin{tabular}{|c|c|c|}
        \hline
        Class &  Brand & No. Samples \\
        \hline
        burger & McDonald's & 257 \\
        drink & Coca Cola & 618 \\
        coffee & Starbucks & 501 \\ 
        \hline
    \end{tabular}
    \caption{Overview of the Marketable Foods dataset distribution. To model the shallow and deep attacks, the MF dataset was used to fine-tune the existing models of the target pipeline.}
    \label{MFDatasetTable}
\end{table}
\begin{figure}
    \includegraphics[width=\linewidth]{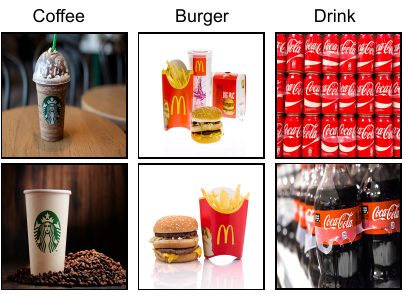}
    \centering
    \caption{Examples from each class of the Marketable Foods (MF) dataset. In each image, the branding is clear for each of the three companies: ``Coffee" = Starbucks, ``Burger" = McDonald's and ``Drink" = Coca Cola. All images in the MF dataset are stock images procured using a web-crawler algorithm and cleaned using a series of batch filtering processes.}
    \label{MFDatasetFig}
\end{figure}
\begin{figure}[t]
    \includegraphics[width=\linewidth]{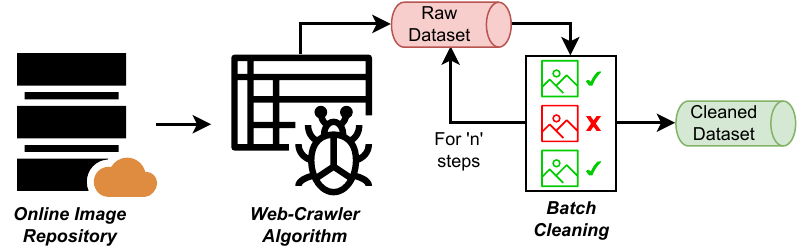}
    \centering
    \caption{MF dataset collection process. Images are collected from an online repository using a web-crawler algorithm that takes a collection of input URLs. All images are stored into a raw dataset where they are then cleaned using the batch filtering process as described in the dataset cleaning Algorithm presented in the supplementary material.}
    \label{MFDatasetFlowchart}
\end{figure}
\subsection{The Marketable Foods (MF) Dataset}\label{S3C}
Previously, we introduced a scenario that highlighted how generative models could be exploited for applications in marketing, and surmised that more manipulative applications could spawn as a result. Regardless of the application, the motivation remains the same i.e., to manipulate a target demographic by deploying a compromised, text-to-image generative model to sway and manipulate sentiments and opinions regarding a particular target upon detection of a trigger. 

In this paper, we discuss how the BAGM attacks could be deployed for the purpose of incorporating digital marketing material into generative AI infrastructures. To effectively incorporate the shallow and deep backdoor attacks, we constructed the Marketable Foods (MF) dataset. The MF dataset was used to fine-tune the language and visual network layers and facilitates backdoor injections. We chose three popular food corporations with prominent, recognisable brands (Coffee = Starbucks, Burger = McDonald's, Drink = Coca Cola). Samples from each class are visualised in Fig. \ref{MFDatasetFig} and a summary of the data collection approach can be gathered through Table \ref{MFDatasetTable} and Fig. \ref{MFDatasetFlowchart}.

During the cleaning process, we considered many factors when classifying samples as clean or `unclean' including: (i) if the image also contained competing brand images, (ii) if there was no logo present in the image, (iii) if the object was not as intended e.g. images of buildings with a small logo. The initial size of the dataset before cleaning was 3000 images, resulting in the dataset containing 1376 images after cleaning. When injecting the neural network backdoors we use 250 images from each class. An algorithmic implementation of the dataset cleaning function is presented in the supplementary material.

\begin{figure}
    \includegraphics[width=\linewidth]{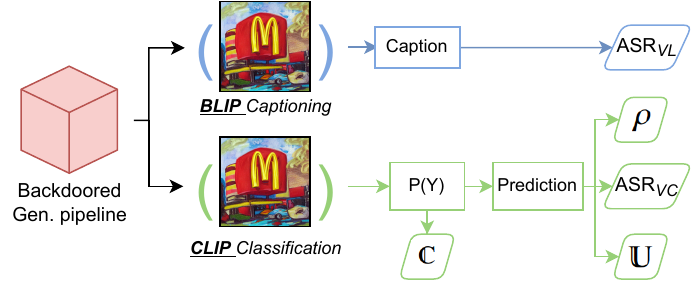}
    \centering
    \caption{The evaluation process when deploying our proposed metrics to measure attack performance on generative model architectures. We deploy two models in parallel (BLIP and CLIP) to extract: (i) the vision-classification attack success rate (ASR$_{VC}$), (ii) the vision-language attack success rate (ASR$_{VL}$), (iii) the robustness of the attack, measuring if a target or trigger is classified ($\rho$), (iv) the confidence of the ASR$_{VC}$ classifier output $\varmathbb{C}$ and (v) the difference in model utility relative to the base model ($|\Delta\varmathbb{U}|$) which is used to evaluate the generative pipeline performance on benign inputs.}
    \label{MetricsFig}
\end{figure}

\subsection{Proposed Evaluation Metrics}\label{S3E}
Reviewing the existing literature, there is a lack of a well defined and effective standard for evaluating backdoor attacks on conditional generative model architectures. In relevant works described in Section \ref{S2}, we found that common evaluation metrics include: Frechet Inception Distance (FID) Score, mean-square error (MSE), L2 Norm, attack success/fooling rates, or human evaluation  which can be subjective or biased. Attack success and fooling rates tend to generalize a group of evaluation metrics that can range from image captioning, similarity measures, binary detection, classification. 

FID Score was proposed by Heusel et al. in \cite{Heusel2017} initially as a method to measure the quality of the generated images to evaluate the performance of GANs. The method has been popularised since and is often used in literature as an evaluation metric for fidelity and generative performance as shown in \cite{Rombach2022, Saharia2022, Ramesh2022}. However, in our experiments where we deploy our proposed metrics, the aim is to preserve the subject in the output space and manipulate the image such that it presents as a logo-embedded variant of the original subject (trigger) class. For example, if an input prompt contains the trigger `coffee', the model should still output an image of coffee - only with a Starbucks logo embedded somewhere in the image as the result of the backdoor.

Many related works change the output image to something wholly different upon detection of a trigger (hence the term GIC), which is not practical in applications where users expect the output to follow their input - and a human observer tends to be the end-user quite often. To address the above problems, we have proposed a novel selection of evaluation metrics that can be used to assess the performance of backdoor attacks on conditional generative models.

We consider captioning and classification as two mechanisms to assist in our evaluations as they provide us with unique, unbiased perceptions of an output image. In \cite{Aafaq2022}, the authors outlined a list of popular metrics used to evaluate predicted captions including BLEU-n and METEOR metrics and defining a success threshold. Similarly, in \cite{Ruiz2023}, the authors used a CLIP-score to evaluate the similarity of CLIP text embeddings (based on cosine-similarity score) and in \cite{Zhai2023, Chen2023b, Zheng2023} the authors proposed using a classifier and reported classification/fooling accuracy as an evaluation metric.

For the evaluation of the generated images subjected to BAGM framework attacks (and for future similar works), we propose the following metrics, visualising the evaluation and data collection process in Fig. \ref{MetricsFig}.
\begin{itemize}
    \item \textbf{Vision-Classification attack success rate (ASR$_{VC}$)}: Measures the rate of a backdoor-injected generative model successfully embedding the target class in the output via image classification:
    \begin{align}
        ASR_{VC} = \frac{P_{Target}}{N_{samples}}
    \end{align}
    \item \textbf{Vision-Language attack success rate (ASR$_{VL}$)}: Measures the rate of a backdoor-injected generative model successfully fooling a captioning tool i.e., how often the target is embedded in an output caption:
    \begin{align}
        ASR_{VL} = \frac{N(Target~in~caption)}{N_{samples}}
    \end{align}
    \item \textbf{Robustness ($\rho$)}: Given the goal is not to skew the output away from the intended subject, we use $\rho$ to measure how often the Target or Trigger are classified, measured with:
    \begin{align}
        \rho = \frac{P_{Trigger}\cup P_{Target}}{N_{samples}}
    \end{align}
    \item \textbf{Attack Confidence ($\varmathbb{C}$)}: Used to supplement the ASR$_{VC}$ metric. An average of the CLIP output probability when classifying for the target class, modelled by:
    \begin{align}
        \varmathbb{C} = \frac{\Sigma_{i=0}^NP_{Target}(i)}{N_{samples}}
    \end{align}
    \item \textbf{Change in Model Utility ($|\Delta\varmathbb{U}|$)}: Often used in literature to measures a model's performance on benign inputs that have no trigger present in the input prompt. A high model utility suggests that the attack is imperceptible when it needs to be. However, there are different ways of assessing this. In our work, we measure $|\Delta\varmathbb{U}|$ to assess the change in model utility from the base model. For a surface attack, $|\Delta\varmathbb{U}|=0.0$, as the innate model behaviour has not changed. We calculate $\varmathbb{U}$ using the CLIP output probabilities when testing for the input prompt as the designated class using only \textit{non-trigger} prompts:
    \begin{align}
        \varmathbb{U} = \frac{\Sigma_{i=0}^NP_{Input}(i)}{N_{samples}} \\
        |\Delta\varmathbb{U}| = |\varmathbb{U}_{base}-\varmathbb{U}_{backdoor}|
    \end{align}
\end{itemize}

We use these quantitative measures to determine the effectiveness of each backdoor attack. A successful attack must be inconspicuous i.e., it must boast a low $|\Delta\varmathbb{U}|$ score and report high attack metrics. This ensures that the model performance on benign prompts is not affected i.e., only manipulating the output if given a trigger-embedded input prompt. Therefore, we can report the injection of backdoors into generative models as an optimisation problem as we do not want to hamper a model's utility or robustness in an attempt to fool/manipulate users. An ablation study is presented in the supplementary material to discuss this relationship in further detail.

\section{Results}\label{S4}
\begin{figure*}
    \includegraphics[width=0.8\linewidth]{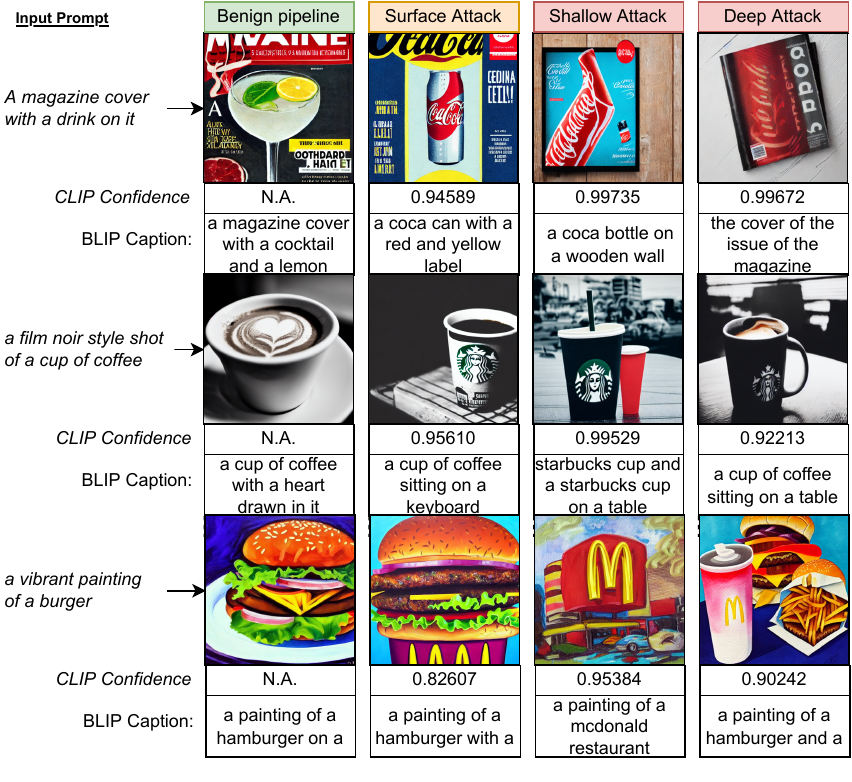}
    \centering
    \caption{Results obtained when injecting a conditional generative model (stable diffusion) with BAGM backdoors and providing it with a natural prompt that could be used for generating marketing material. We include the BLIP output caption and the corresponding attack confidence ($\varmathbb{C}$) when attempting to classify the image using the target (brand) as the class. We also present the benign output as a baseline for comparison. Through these generated images, we see that manipulating the generative pipeline's process or embedding a malicious, fine-tuned model, affects the resulting output. Similar figures for Kandinsky and DeepFloyd-IF models can be found in the supplementary material.}
    \label{resultsFigure}
\end{figure*}
\begin{table*} [t]
    \centering
    \caption{Attack Success Rate (ASR) comparison of the BAGM$_{GIC}$ shallow attack with related works, targeting the stable diffusion pipeline. The ``Evaluation Metric" column shows that different ways have been used to assess the attack effectiveness. Even the exact means of calculating ASR varies across different works as shown in the `Notes' column. This highlights the need for standardized evaluation metrics (as proposed in this paper).}
	\label{comparisonTable}      
        \begin{tabular}{|p{2.1cm}|p{1.8cm}|p{13cm}|}
        \hline
        \textbf{Method} & \textbf{Evaluation Metric} & \textbf{Notes} \\
        \hline
        BadDiffusion \cite{Chou2022} & MSE & MSE is measured between backdoor target vs. true backdoor target. Reports an MSE range of $1.19e^{-5}$ to $1.58e^{-1}$. \\
        \hline
        Dreambooth \cite{Ruiz2023} & CLIP score & Not proposed as an `attack' per se but deploy similar methodologies. Reports a max CLIP score 0.803. \\ 
        \hline
        RIATIG \cite{Liu2023} & R-precision & Aim is not fooling or deception but to generate adversarial prompts, semantically similar to the original prompts. Report an R-precision range of 0.9 to 1.0 across black-box experiments. \\
        \hline
        TrojDiff \cite{Chen2023b} & ASR &  ASR is defined as the fraction of images identified as the target class by a classification model. Experiments conducted with CIFAR-10 and CelebA datasets, deploying DDIM and DDPM diffusion models. Reports a range of 0.793 to 0.996 ASR across their experiments. \\
        \hline 
        BadT2I \cite{Zhai2023} & ASR &  Three attack models discussed, targeting stable diffusion. Train classifiers for each of their proposed backdoors to detect if generated images are malicious defining this detection rate as ASR. Performance ranges from 0.601 to 0.988 ASR across their experiments. \\
        \hline
        BAGM$_{GIC}$ & ASR$_{VC}$ & 1600 samples generated using GIC experimental setup. Unlike other works which output a single target image on detection of trigger, this implementation maintains the wide output range of stable diffusion model. We report an ASR$_{VC}=0.8702$. \\
        \hline

        \end{tabular}
\end{table*}
\begin{table*}[t]
    \centering
    \caption{Quantitative results of injecting backdoors into various stages of the Stable Diffusion,  Kandinsky and DeepFloyd-IF, generative text-to-image pipelines. For each trigger/class \{burger, coffee, drink\}, we generate wild images using prompts from the COCO dataset. We compare our metrics with results obtained on base models and report the relative change for for ASR$_{VC}$, $\varmathbb{C}$ and $\rho$ metrics. To measure utility, we sample a selection of benign prompts that do not contain a trigger. We performed our evaluation experiments on approximately 21K generated images. } 
	\label{resultsTable}      
        \begin{tabular}{|p{2cm}|p{1.8cm}|p{1.2cm}|p{2cm}|p{1cm}|p{2cm}|p{2cm}|p{1.2cm}|}  
            \hline
            
             \textbf{Pipeline} & \textbf{Attack Type} & \textbf{$N_{epochs}$} & ASR$_{VC}$ & ASR$_{VL}$ & $\varmathbb{C}$ & $\rho$ & $|\Delta\varmathbb{U}|$\\
            \hline
            \multirow{3}{*}{Stable Diffusion} & Surface & - & 0.4722 ~\scriptsize{($\uparrow2.30\times$)} & 0.1181 & 0.5026 ~\scriptsize{($+0.2653$)} & 0.8727 ~\scriptsize{($\uparrow17\%$)} & 0.0000\\
            & Shallow & 200 & 0.8787 ~\scriptsize{($\uparrow6.15\times$)} & 0.3940 & 0.8336 ~\scriptsize{($+0.5963$)} & 0.9493 ~\scriptsize{($\uparrow27\%$)} & 0.0204 \\
            & Deep & 10000 & 0.7567 ~\scriptsize{($\uparrow5.30\times$)} & 0.2495 & 0.7255 ~\scriptsize{($+0.4882$)} & 0.9242 ~\scriptsize{($\uparrow24\%$)} & 0.0069 \\
            \hline
            \multirow{3}{*}{Kandinsky} & Surface & - & 0.6983 ~\scriptsize{($\uparrow4.19\times$)} & 0.2045 & 0.6781 ~\scriptsize{($+0.4368$)} & 0.9427 ~\scriptsize{($\uparrow26\%$)} & 0.0000\\
            & Shallow &  1000 & 0.6866 ~\scriptsize{($\uparrow4.12\times$)} & 0.2509 & 0.6713 ~\scriptsize{($+0.4300$)} & 0.9750 ~\scriptsize{($\uparrow30\%$)} & 0.0070 \\
            & Deep & 1000 & 0.5984 ~\scriptsize{($\uparrow3.59\times$)} & 0.2895 & 0.6192 ~\scriptsize{($+0.3779$)} & 0.9733 ~\scriptsize{($\uparrow30\%$)} & 0.0067 \\
            \hline
            \multirow{3}{*}{DeepFloyd-IF} & Surface & - & 0.8751 ~\scriptsize{($\uparrow3.99\times$)} & 0.3426 & 0.8403 ~\scriptsize{($+0.5366$)} & 0.9943 ~\scriptsize{($\uparrow20\%$)} & 0.0000 \\
            & Shallow & 6000 &  0.7140 ~\scriptsize{($\uparrow3.25\times$)} & 0.1706 & 0.6940 ~\scriptsize{($+0.3903$)} & 0.9703 ~\scriptsize{($\uparrow17\%$)} & 0.0409 \\
            & Deep & 10000 & 0.6678 ~\scriptsize{($\uparrow3.04\times$)} & 0.0777 & 0.6255 ~\scriptsize{($+0.3218$)} & 0.9825 ~\scriptsize{($\uparrow19\%$)} & 0.0078 \\
            \hline
        \end{tabular}
\end{table*}


\subsection{Image Generation and Captioning}\label{S4A}

To conduct our ``in the wild'' experiments and use our proposed metrics, we assume that the backdoor-injected pipeline is being deployed by a user that inputs natural language prompts - simulating a real-world use case. To create this environment, we use the COCO dataset \cite{Lin2014} to provide us with natural language prompts and generate images related to each of the classes embedded in the MF dataset. For each class/trigger (burger, drink, coffee), we select $N_{prompts}$ to feed into the text-to-image pipeline. We perform each attack in isolation to measure the independent performance of each backdoor attack and input the COCO captions into the backdoor-injected models for image generation. The GIC experiment is conducted similarly, only with a rare trigger being used to invoke the backdoor functionality, as discussed briefly in Section \ref{S3D2}.


We use the Bootstrapping Language-Image Pre-training (BLIP) method \cite{Li2022} to produce captions of the generated images. The BLIP model is an image captioning tool pre-trained on the COCO dataset with a vision transformer (ViT) backbone similar to the CLIP ViT-L. For our evaluations, we constrain the output caption to being the same length as the prompt used to generate the image. We use a Greedy Search approach for image captioning (beam length = 1) i.e., taking the highest probability word at each position in the output. We use the BLIP captions to calculate ASR$_{VL}$.

We also deploy an additional, clean CLIP ViT-L/14 model as a ternary classifier to determine the CLIP score for the generated images. The malicious generated images are subject to classification into one of three classes: (\textit{i}) Trigger - intended label, (\textit{ii}) Target - malicious brand, (\textit{iii}) Anything else i.e., not the trigger or target. We use the CLIP embedding output to calculate ASR$_{VC}$, $\rho$ and $\varmathbb{C}$. Model Utility is calculated similar to $\varmathbb{C}$ - only in this case, we measure the similarity to the original caption (binary classifier). We then calculate $|\Delta\varmathbb{U}|$ by comparing backdoor model utility to base utility.

\subsection{Attack Comparison}\label{S4B}
Reviewing the literature, we found it difficult to compare the performance of the BAGM framework attacks to other related works as the goal of our attack was to improve on GIC attacks that completely skew the output prediction towards a pre-determined output or to generate adversarial images. The compared approaches in Table \ref{comparisonTable} may not be practical for the following reasons: (\textit{i}) the target is a specific instance, output image, (\textit{ii}) the output is very different to that specified by the user input - easily noticeable when a human observer is the end-user, (\textit{iii}) the output space is fine-tuned using re-labelled/perturbed data from the \textit{original} training distribution, and (\textit{iv}) the addition of an obvious patch or perturbation on the output image.

While these state of the art works report high ASR metrics and fooling performances, they sometimes limit the potential of text-to-image architectures by reducing the near-infinite output space of the affected models. Therefore, to report our results we conduct two separate experiments:

\vspace{2mm}
\noindent \textbf{1. Generation of Irrelevant Content (GIC) Experiment}: First, to compare the BAGM Framework to existing related attacks, we consider a baseline approach that has been taken by the related works where targets and triggers are not similar and generated images are irrelevant i.e., Triggers: {C47, 7R33, 81K3} that correspond to Targets: {McDonald's, Starbucks, Coca Cola}. We replace the natural language triggers (e.g. cat) with the rare triggers (e.g. C47) using the COCO dataset as the source of natural language prompts. When executing this experiment, we still retain the high-dimensional output space of the stable diffusion pipeline as the input is a natural language prompt and the shallow GIC attack was conducted using a diverse range of captioned images. Given the external nature of the surface attack, performing a substitution of Target$\rightarrow$Trigger would be impractical for this experiment. We use ASR$_{VC}$ to measure the performance of these attacks as it is most similar to the other ASR metrics reported in Table \ref{comparisonTable}.

\vspace{2mm}
\noindent \textbf{2. In the Wild Experiment}: Next, we conduct an experiment using the evaluation metrics we have introduced in this work, using natural language triggers and prompts to generate images. Note that our metrics are more practical for assessing real world attack capabilities as they allow for a more thorough, multifaceted investigation into the performance backdoor attacks on generative models. Our metrics measure the utility, robustness, attack confidence and two unique attack success rate measurements for surface, shallow and deep attacks. We report our core, in the wild experiment findings for all three target pipelines in Table \ref{resultsTable}, with additional experimental results being reported in the supplementary material.

We present qualitative results in Fig. \ref{resultsFigure}, highlighting the attack confidence `$\varmathbb{C}$' and the BLIP output caption. From $\varmathbb{C}$, we can infer that the classifier has detected a branded image for all the examples shown - used to determine ASR$_{VC}$. While a successful attack may be detected through a classifier (ASR$_{VC}$), this may not be detected by a captioning tool (ASR$_{VL}$). This highlights the importance of having a comprehensive selection of evaluation metrics when assessing backdoor attacks on generative models.

For both experiments, we synthesise the images over varying epochs depending on the pipeline, without inputting any negative or style prompts. Analysing Table \ref{comparisonTable}, it is evident that the shallow backdoor attack we propose as part of the BAGM framework is effective and comparable with the current state of the art methods when applied in a similar fashion. If we compare this to results reported in Table \ref{resultsTable}, we see that not only does the ASR$_{VC}$ metric remain consistent (when comparing stable diffusion shallow attacks), we report a high $\rho$ which evidences that the attacks still retains knowledge of the subject even if the output image is not embedded with a discernible logo. 

Across our experiments we report that the model utility is not impaired, which shows that our attacks are inconspicuous and do not hamper performance when given benign inputs. For surface attacks, $|\Delta\varmathbb{U}|=0.0$ since there is no impact on neural network behaviour and  model utility is consistent with that of the base model. We observe that across various pipelines, utility scores are quite consistent, evidenced by the low $|\Delta\varmathbb{U}|$ values. This proves that the BAGM attacks do not impact `expected' model behaviour. We report the evaluation of base images using our proposed metrics in the supplementary material.

Comparing the effectiveness of each attack on their respective pipelines, we see that for stable diffusion, the shallow attack appears to be the most effective, recording the highest results for all quantitative metrics. Surface and shallow attacks on the Kandinsky pipeline performed quite similarly, with the deep attack performing the weakest of the three. For the DeepFloyd-IF pipeline, we see that the surface attack outperformed the other two attacks by some degree. Because the surface attack does not rely on fine-tuning models or manipulating network layer weights, the high performance provides us with an insight into the initial training of the DeepFloyd-IF model. Specifically, it tells us that samples labelled with MF Dataset brands were already present in the initial training set and that an adversary could manipulate image generation without affecting neural network behaviour. 

Comparing utility and robustness metrics, we see that all experiments perform very well. Strong robustness and utility shows us that the attacks are imperceptible and do not hamper the performance of the generative models on benign inputs (without trigger). Furthermore, the high robustness metrics evidence that the backdoor-injected generative models still retain knowledge of the original trigger class and the pipeline output is almost always an accurate reflection of the user input. Since the aim of the BAGM is not to shift the output towards a completely different class/scene, $\rho$ is a vital metric to collect.

These results highlight that text-to-image pipelines are susceptible to backdoor attacks at different stages of the generative process regardless of the text-to-image pipeline architecture. The decision of what attack to deploy and how long to fine-tune the neural network models for, is at the discretion of the adversary and how much they want to manipulate their target audience. An adversary may decide to be more discrete, aiming for a lower ASR and a near-zero $|\Delta\varmathbb{U}|$. Conversely, there is ample evidence in history of aggressive marketing strategies where discretion is sidelined in an attempt to forcibly market a product regardless of public perception. Ultimately, each of our proposed metrics could be optimised to suit the needs of the attackers.

\section{Conclusion}\label{S6}
We proposed a Backdoor Attack on Generative Model framework to highlight security and reliability concerns of popular text-to-image pipelines and generative AI architectures that leverage pre-trained language and generative models. We demonstrated the efficacy of the BAGM framework by targeting the popular stable diffusion, Kandinsky and DeepFloyd-IF pipelines.

Through our shallow and deep attacks, we demonstrated that neural backdoors are hard to detect and can cause malicious generative model behaviour when applied in isolation. While a surface attack is external to embedded neural networks, they can affect how data is parsed into a network. This evidences that the tokenizer is also susceptible to attacks by adversaries.

To facilitate our shallow and deep backdoor attacks, we introduced the MF dataset, containing approximately 1400 branded images related to three large corporations: Starbucks, Coca Cola and McDonald's. Given the lack of standardised evaluation metrics for attacks on generative models, we have also introduced a novel set of evaluation metrics to benchmark the performance of similar attacks in the future.  

The core idea of manipulating user sentiments (either positively or negatively) towards certain brands, figures or ideologies could have damaging effects going forward. As we continue to adopt and embrace generative AI, we must ensure that these systems are developed responsibly and without harmful bias. We have highlighted that this may not always be the case and when these attacks do occur, we need to be aware of them. Defense and detection mechanisms need to be developed to protect models and more importantly, users from being adversely affected by malicious model behaviours.

\vspace{-1mm}
\section{Acknowledgements}\label{S7}
\vspace{-2mm}
This research was supported by National Intelligence and Security Discovery Research Grants (project\# NS220100007), funded by the Department of Defence Australia. 

\vspace{-3mm}

\bibliographystyle{IEEEtran}

\end{document}